# Sentence based Semantic Similarity Measure for Blog-Posts

Mehwish Aziz

Computer Science Department
National University of Computer & Emerging Sciences,
Karachi Campus
mehwish.aziz@nu.edu.pk

Muhammad Rafi

Computer Science Department
National University of Computer & Emerging Sciences,
Karachi Campus
muhammad.rafi@nu.edu.pk

*Abstract*— **Blogs-Online digital diary like application on web 2.0 has opened new and easy way to voice opinion, thoughts, and like-dislike of every Internet user to the World. Blogosphere has no doubt the largest user-generated content repository full of knowledge. The potential of this knowledge is still to be explored. Knowledge discovery from this new genre is quite difficult and challenging as it is totally different from other popular genre of web-applications like World Wide Web (WWW). Blog-posts unlike web documents are small in size, thus lack in context and contain relaxed grammatical structures. Hence, standard text similarity measure fails to provide good results. In this paper, specialized requirements for comparing a pair of blog-posts is thoroughly investigated. Based on this we proposed a novel algorithm for sentence oriented semantic similarity measure of a pair of blog-posts. We applied this algorithm on a subset of political blogosphere of Pakistan, to cluster the blogs on different issues of political realm and to identify the influential bloggers.**

## I. INTRODUCTION

Blogs-online digital diary like application on World Wide Web (www) opens new horizons for all internet users to voice their opinions, thoughts, and likes-dislikes to the entire world. The rapid popularity of blogging is due to fast, easy and effective content creation, collaboration and dissemination. The potential of this users' generated knowledge repository is theoretically unlimited. Recently, researchers introduced new approaches for opinion extraction, community identification, clustering blog-posts, and identifying influential bloggers from the blogosphere. They also highlighted that the traditional text mining methods are not well suited for this new genre of web-application. The careful analysis of blog-post vs. web-document reveals that the blog-post requires special consideration due to its special post structures, short length, lack of context and relaxed grammars [1]. It is often required to compare a pair of posts for various types of processing; one such application is to cluster the blog-post on certain topics. Similarity measure is critical in determining the relatedness between a pair of blog-posts so an effective clustering can be done. The traditional similarity measures of text are based on term matching, which fails to capture the semantic of the text. Several efforts have been made to define semantic similarity measures for text; most of them determine the language semantic through word annotations and relating the concept space through replacement of equivalent concepts by using a lexical database like WordNet. It is often difficult to build and determine context of a short segment of text through these approaches. We suggest here a novel approach, which annotates sentence through part-of-speech taggers, then we collect noun phrases as it usually represents the common semantic of the text from the pair of blog-posts. Our similarity measure utilizes weighted similarity measure of noun phrases, verb phrases and the common bag- of- words, hence it is generalized enough to capture true semantic of the blog-post. The experiment performed on the dataset clearly exhibited that the similarity measure defined in this paper is practicable.

In the next section we will discuss the related work from our fellow researchers, followed by description and discussion of our approach, and thereafter, the experimental data and set-up will be discussed followed by the illustration of the result of the experiment. This paper concludes with a discussion on future extension of this work along with the possible application of the suggested algorithms on blog-posts similarity measure based on sentence-semantic.

## II. LITERATURE REVIEW

Web 2.0 has brought interactive and easy mechanism for user generated knowledge contents. Blogs is one genre that is growing in popularity and application. It is the largest user generated knowledge repository on the web. The potential for this knowledge is still to be explored by both individual and corporate users of Internet, who are slowly discovering the business value of it. Researchers recently inclined towards knowledge discovery in the blogs [1], particularly identifying and mining sentiments or opinions [2], [3] and [4], identifying influential bloggers [5] and [6], community identification and clustering of blog-posts [7], [8] and [9]. A thorough analysis of blogosphere demands that the traditional metrics of information mining must be redefined for the blogosphere. An information measure that is widely used in information retrieval and mining is similarity measure. Similarity measure is a function that gives a numerical value to a pair of entities; this numerical value represents the similarity between them. Higher numerical value demonstrates high similarity between entities like; pair of documents, images, products, and services. This similarity measure can be used in a variety of scenarios where the sense-of-relatedness is used for evaluation. One such example is to cluster the blog-posts on some given topic. Approaches to text similarity measure [10] can be subdivided into two broad categories (i) Dictionary/Thesaurus based and (ii) information theoretic based (corpus based). The dictionary based approach is particularly useful with highly constrained taxonomies like domain specific semantic net, a detailed discussion can be found in [11]. One major drawback of this approach is that it only considers IS-A relationships by considering Hyponymy/ Hypernymy. It does not include antonymy, and part-whole relationship. The other approach based on information theory is introduced by Resnik at. el. [12], which defines the similarity between two concepts as a maximum information content related to concept that subsumes them in the taxonomy hierarchy. A combined approach like our approach is first introduced by Jiang and Conrath [13], it define the similarity between two concepts as the path based information contents that maximize the semantic of the concept. This is computed as the weighted-edge path along the shortest path linking the two concepts.

## III. BLOG-POSTS SIMILARITY MEASURE

The semantic similarities of two blog-posts mean that the knowledge contents of the two posts are similar. Thus the concepts that the two posts discuss are relevant. The short text, grammar-free and lack of context are the challenging issues for computing semantic similarity of blog-posts. We proposed a novel algorithm named "Blog-Post Similarity Measure" i.e. BPSM. This algorithm is evaluated over all the pairs of distinct blog posts in the dataset. In order to capture the semantically true similarity measure of the blog posts, the algorithm performs certain steps which are explained below.

At first the algorithm performs iteration over entire dataset so as to pair-up all distinct blog posts (say $Pk$ and $Pl$) to measure sentence-based similarity. The next step is to utilize the sentence-based parts of speech taggers for identifying distinct set of noun phrase, verb phrase and common bag–of-words of blog posts $Pk$ and $Pl$. Following these steps, the next step would identify the set of similar noun phrases, verb phrases and common bag-of-words from the distinct sets of both the blog posts $Pk$ and $Pl$ named as sim_noun, sim_verb and sim_common respectively. In order to capture the semantically true similarity, convex combination (variation of weighted mean) is calculated. For convex combination , weights are assigned for sim_noun, sim_verb, sim_common named as alpha ($\alpha$), beta ($\beta$) , gamma ($\gamma$) using equation (1), (2) and (3) respectively.

$$= (sim\_noun / (sim\_verb + sim\_common)) \quad (1)$$
$$= (sim\_verb / (sim\_noun + sim\_common)) \quad (2)$$
$$= (sim\_common / (sim\_verb + sim\_noun)) \quad (3)$$

Finally, the algorithm utilizes these weighted similarity measures of noun phrases, verb phrases and the common bag-of-words to calculate the similarity measure between blog posts $Pk$ and $Pl$ using convex combination in equation (4) shown below.

$$SIM\_Measure = ((( * sim\_noun) + ( * sim\_verb) + ( * sim\_common)) / total\_word\_list) \quad (4)$$

Here, total_word_list indicates the accumulation of sim_noun, sim_verb and sim_common as shown in equation (5).

$$total\_word\_list = (sim\_noun + sim\_verb + sim\_common) \quad (5)$$

The similarity measure is mainly using noun and verb weights for actual semantic extraction from the sentence whereas common bag-of-words' weights are used due to the probabilistic nature of the language.

## IV. EXPERIMENT

The suggested algorithm is used to perform an experiment on a sub-set of political blogosphere, from the Pakistani political blogs. As the purpose of this paper is to show the sentence-based similarity measure, therefore, we gathered the blog posts of a particular category "Pakistani Politics" from RSSFeeds generated by different blog sites. For analyzing the performance of the algorithm, an offline data set of 2700 Blog Posts is collected from different Blogs over "Pakistani Politics" in the time duration of Jul, 2007 till Feb, 2010. As the Blog Posts' data set is huge and it becomes cumbersome to measure the efficiency of the algorithm over Blog Posts holding variant political issues within Pakistani politics. The data set is further categorized from "Pakistani Politics" into the sub-categories named as "Political Issues" which are stated below:

0. Politicians
1. Judiciary
2. Musharraf
3. Terrorism
4. Taliban
5. War on Terror

Similarity measure over the Blog Posts is performed using the entire dataset but a partial subset of 60 Blog

Posts relevant to the "Political Issue" stated above is extracted by performing soft clustering to better understand the algorithm results. Thus, only 10 Blog Posts relevant to each political issue is assigned in a cluster. A single blog-post can be assigned to more than one cluster due to the content relevancy in different blog-posts. The Blog Posts (identified by their unique IDs) and their distribution over the political issues are shown below in the figure 1.

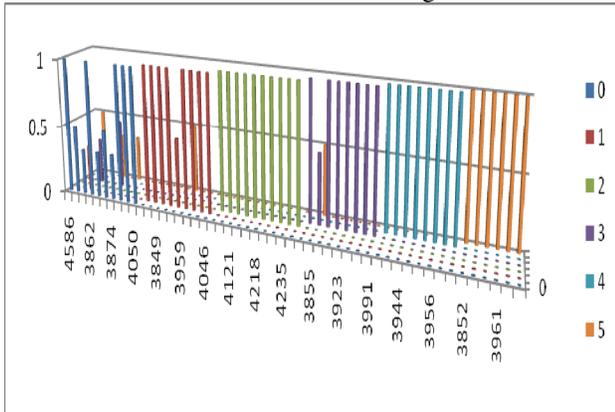

**Figure 1:** Distribution of 60 Blog Posts over the political issues. Here Political Issues are identified as 0 = Politicians, 1 = Judiciary, 2 = Musharaf, 3 = Terrorism, 4 = Taliban, 5 = War on Terror. And Blog Posts are identified over the range of Post IDs from 3800 to 5500 on x-axis. 0 to 1 on y-axis shows the similarity measure value for corresponding Post IDs.

In order to help in understanding the results obtained from the testing dataset defined in Figure 1, the testing data set is further summarized by selecting only 20 Blog Posts i.e. at least 3 unique Blog Posts from each political issue. Now lets' review the 20 Blog Posts' distribution in Figure 2.

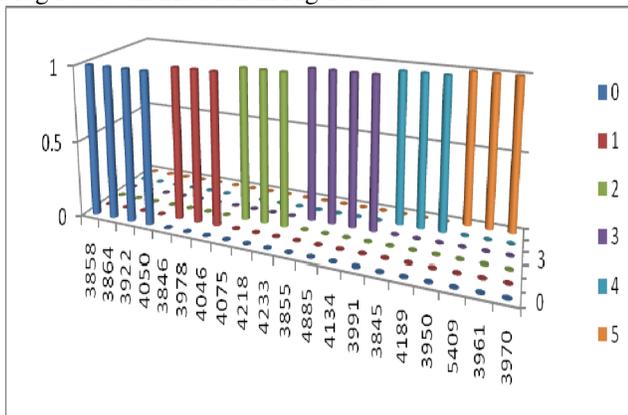

**Figure 2:** Distribution of 20 Blog Posts over the Political Issues (subset of original testing dataset). Here Political Issues are identified as 0 = Politicians, 1 = Judiciary, 2 = Musharaf, 3 = Terrorism, 4 = Taliban, 5 = War on Terror. And Blog Posts are identified over the range of Post IDs from 3800 to 5500 on x-axis. 0 to 1 on y-axis shows similarity measure value for corresponding Post IDs.

All these blog posts collected as an offline data set are stored in a relational database in an XML-Format shown in Figure 3. These blog posts' texts comprise of both English and Roman Language based words.

```
<ParsedRSSFeeds>
<ParseRSSFeed_ID>4586</ParseRSSFeed_ID>
<ID>3941</ID>
<Title>Zardari Fights for Survival Following SC Verdict on NRO</Title>
<Desc>PKonweb Monitorln a major blow to President Asif Ali Zardari, the Supreme Court today unanimously declared the controversial National Reconciliation Ordinance (NRO), which provided amnesty to political leaders against corruption charges, an 'unconstitutional' and against 'national interest.'The verdict by the full (17-member) bench of the apex court headed by Chief Justice Iftikhar Muhammad Chaudhury is likely to stir up more turmoil in the country weakening the Presidency further.The populist landmark historical decision by the newly independent judiciary has been hailed by the civil society and the common man on the street including the media.President Zardari no longer has the moral authority and ground to head the federation, many say. But, whether he will do so or chose to seek cover behind the cloak of immunity guaranteed to Presidents under the constitution is another question.Petitions challenging Zardari's eligibility as a presidential candidate are expected to follow from the court's ruling. and about a dozen senior members of Zardari's coterie of advisers will most likely face renewed corruption cases.However, Presidential spokesperson, Farhatullah Babar, stressed both Zardari and the Pakistan People's Party (PPP) respect the court's verdict, and that it would not affect the immunity enjoyed by the President."We believe that no criminal case can be instituted or continued in any court against a president or a governor during the term of office. So, this doesn't affect the president of Pakistan. Regarding other matters, the law will take its course and we will see what happens," the newspaper quoted Babar, as saying.Zardari's close aide, Dr. Babar Awan, too rebuffed opposition's call for President's resignation, saying he would complete his term in office.During the hearing, which was closely followed and scrutinized by the local media, the court raised objections, in particular, over the question of who had authorized the return of 60 million dollars in suspect gains by Zardari to foreign companies in his name after the government withdrew criminal proceedings against him in Switzerland last year.The Supreme Court , in its verdict, said that the withdrawal of the cases against Zardari in Switzerland, which was ordered by the former Attorney General, Malik Qayyum, was illegal and that the government should contact the Swiss authorities to restore the proceedings. The court also asked the government to punish Qayyum for his unauthorized act including replacing the present NAB Chairman, the prosecutor and his deputy on grounds of lack of performance and mistrust.While the repercussion of the apex court's monumental judgment is yet to be seen, but it has certainly added fuel to the country's troubles, which is already fighting for its existential survival under the barrage of terror attacks being carried out by extremists.<div class='blogger-post-footer'>Pakistan Politics, Best source for quality articles on South Asiapakistanpal.blogspot.com' alt=' '></div><</Data>
<BloggerNoreply@blogger.com'Pakistan Pal</Blogger>
<Pub_Date>Fri, 18 Dec 2009 08:02:00 +0000</Pub_Date>
</ParsedRSSfeeds>
```

**Figure 3:** Example of an XML-Format Blog Post received through RSSFeeds.

First, these blog posts are parsed into sentences by extracting the blog-post content from XML-Format. Then the sentences of blog-posts are further extracted into words with their parts of speech, synonym list (synset) and words' origin/category using hypernym relationship by using natural language processing functionalities in nltk-python by the help of WordNet knowledge base. All these extracted sentences and words are stored post-wise in a relational database – a snapshot of which is shown below in Figure 4.

| ParseRSSFeed_ID | ID | Title | Data | Blogger | Pub_Date | Comments_Count | Category |
|---|---|---|---|---|---|---|---|
| 3848 | 82 | SEO Analysis or | This is a generi | noreply@blog | Sun, 12 Apr 20C | | latest news |
| 3849 | 82 | Restoration of | Struggle of pak | noreply@blog | Thu, 19 Mar 20 | | Civil-Society-P |

| Sentence_IE | Sentence |
|---|---|
| 403 | Restoration of judiciary or independent judiciary |
| 404 | Struggle of pakistani nation has come to an end on the decision of 16 March by Prime Minister Yousuf Raza Gilani to |
| 405 | Lawyers, political activists and civil society celebrated the day as a new opening of justise & equality in the country |

| Word_ID | Word | Parts_of_Sp | Word_Category | Word_Synor_List |
|---|---|---|---|---|
| 2550 | Lawyers | Noun + plural | professional 'professional_per | N/A |
| 2551 | political | Adjective | N/A | 'political' 'political'' |
| 2552 | activists | Noun + plural | reformer 'reformist' 'crusader' | N/A |
| 2553 | and | Coordinating \ | N/A | N/A |
| 2554 | civil | Noun + singu | N/A | civil' 'polite' 'civil' 'civic' 'civil' 'civil' |
| 2555 | society | Noun + singu | social_group' | club' 'social_club' 'society' 'guild' 'gild' 'lodge' 'order' 'o |
| 2556 | celebrated | Verb+past_tn | N/A | celebrate 'fete' 'lionize' 'lionise' 'celebrate' 'celebrate' |
| 2557 | day | Noun + singu | time_unit 'unit_of_time' | day' 'day' 'daytime' 'daylight' 'day' 'day' 'side |
| 2558 | as | Preposition/c | chemical_element 'element' | American_Samoa 'Eastern_Samoa 'AS' equally 'as' es |
| 2559 | new | Adjective | N/A | fresh 'new' 'novel' 'raw' 'new' 'new' 'unexampled' new |
| 2560 | opening | Verb+gerunc | scope' | opening 'opening' 'opening' 'opening_night' 'curtain_r |
| 2561 | of | Preposition/s | N/A | N/A |
| 2562 | justise | Noun + singu | | N/A |
| 2563 | equality | Noun+singu | sameness' | equality 'equivalence' 'equation' 'par' |
| 2564 | in | Preposition/s | linear_unit 'linear_measure' | indium 'In' atomic_number_49 'Indiana 'Hoosier_Star |
| 2565 | country | Noun + singu | N/A | N/A |
| * (New) | | | | |

| | |
|---|---|
| 406 | It will be a check on the Government temper which help stopping them to take strange decisions, with a hope that it |
| 407 | googleusercontent |
| 408 | com/tracker/91662719355333024834-12097788813999110637?incapitaltalk |

Record: I ◀ 1 of 16 ▶ ▶I ▶*   Search

**Figure 4:** Blog Post stored with all the sentences where each sentence's words are stored with their parts of speech, category and synonym list.

Once the blog-posts are processed using natural language processing task by the help of WordNet, then this testing dataset is ready to evaluate the algorithm (BPSM) devised for sentence-based semantic similarity measure between the blog-posts. The algorithm (BPSM) used for such similarity measure is given below.

## V. BPSM-ALGORITHM

**Algorithm – Blog Post Similarity Measure (BPSM)**

Step 1: For each post Pk where k = 1 to n do

1.1) For each post Pl where Pl = 1 to n ∧ Pk ≠ Pl do

i) For Pk and Pl, calculate *total count of sentences* in blog post as count_sentences respectively

ii) Move to Step 2 and identify distinct set of noun, verb and common bag-of-words as nounlist, verblist, commonlist respectively

iii) Calculate *sum of similar nouns, similar verbs and similar common bag-of-words* as sim_noun, sim_verb, and sim_common respectively, as follows:

a) For each noun $n_i$ in Pk where i = 1 to count_noun do
   For each noun $n_j$ in Pl where j= 1 to count_noun do
   - if ($n_i$ equals to $n_j$) or ($n_i$'s synonym list contains $n_j$) or ($n_i$'s category contains $n_j$) then add 1 in sim_noun

b) Do the same as Step 1 – iii – a for both verbs and common bag-of-words and store the result in sim_verb and sim_common respectively

iv) Calculate total_word_list, α(for sim_noun), β (for sim_verb) and γ (for sim_common) using equation (5), (1), (2) and (3) respectively

v) Calculate Similarity Measure for Pk and Pl by using equation (4)

Step 2: For a Post - identify the distinct nounlist, verblist and commonlist

2.1) For each Sentence Si where i = 1 to count_sentences

i) Update nounlist, verblist and commonlist by adding all words/phrases of Si if:

   a) Parts_of_Speech is either of the following: Proper_noun + singular / Noun + singular_or_mass / Noun + plural. Then update nounlist and add 1 in count_noun

   b) Parts_of_Speech is either of the following: Verb + non-3rd_ps.sing.present/ Verb + gerund/present_participle/ Verb + gerund/present_participle/ Verb + base_form/ Verb + past_tense. Then update verblist and add 1 in count_verb

   c) Parts_of_Speech matches neither of above. Update commonlist, add 1 in count_common

2.2) Update nounlist: for each noun $n_i$ where i = 1 to count_noun do

i) for each noun $n_j$ where j = i+1 to count_noun do

   - if ($n_i$ is equals to $n_j$) or ($n_i$'s synonym list contains $n_j$) or ($n_i$'s category contains $n_j$ ) then remove nj and update count_noun

2.3) Do the same as Step 2.2 for both verblist and commonlist. Then move to Step 1-iii

For better understanding of the algorithm (BPSM), its processing is shown as flowcharts. The first flowchart in Figure 5 shows the flow of measuring similarity between blog-posts of the same category. Figure 6 also shows the process of measuring similarity between blog posts Pk and Pl.

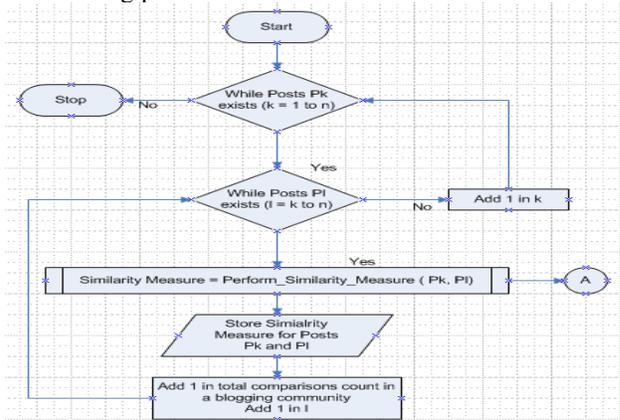

**Figure 5:** Process of measuring similarity between blog-posts of a particular category

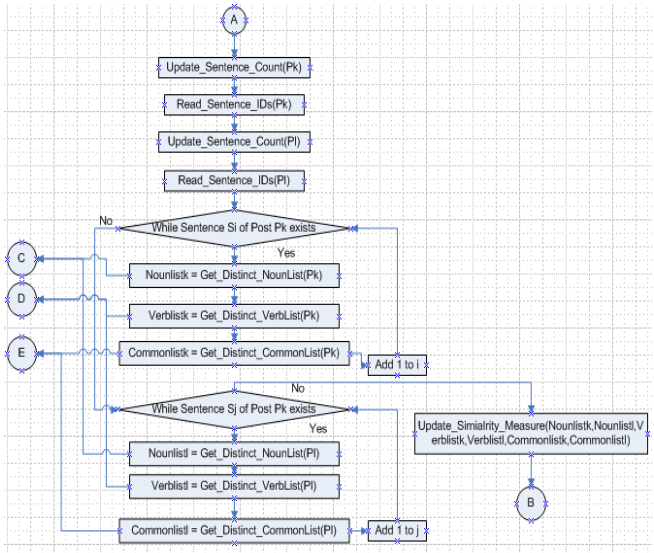

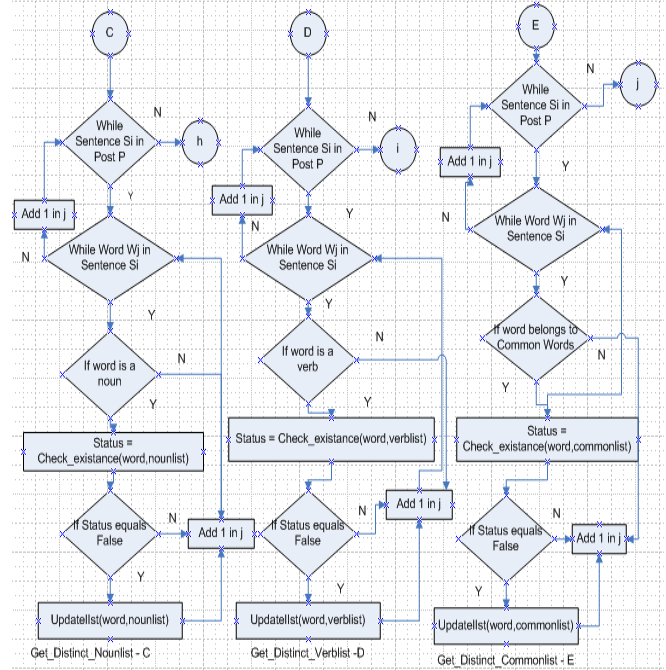

**Figure 6:** Process of measuring sentence-based similarity between two posts Pk and Pl

This sentence-based similarity measure between blog posts in a category shown in Figure 5 and Figure 6 is using the words on the basis of WordNet knowledge base as per their existence in the sentences post-wise. This means that instead of matching two words as terms; the words in the sentences of two posts are matched to update the similarity measure. This update is possible only if the words of sentences in the posts satisfy any of the following criteria:

- Word Wi existent in the sentence Sj of post Pk is matched exactly with Word Wr in the sentence Sm of the post Pl
- Word Wi's synonym list has a word that matches exactly with Word Wr in the sentence Sm of the post Pl
- Word Wi's origin entity matches exactly with Word Wr in the Sentence Sm of the Post Pl

In the flowchart, post Pk is iterated over all the blog-posts in the blogging community, where Sj denotes $j^{th}$ sentence iterated from 1 till the total number of sentences in the post Pk. Similarly, Wi is referring to $i^{th}$ word from 1 till the total number of words in the sentence Sj of post Pk.

Now for each sentence Sj of Post Pk, all the sentences Sm of post Pl are traversed to measure the similarity in such a way that for each distinct word Wi of Sentences Sj of Post Pk is matched for similarity measure with all the distinct words of all the sentences Sm of post Pl. Figure 7 and Figure 8 given below explains the above stated criteria to pdate similarity measure for two blog posts.

**Figure 7:** Create distinct noun list, verb list and common list

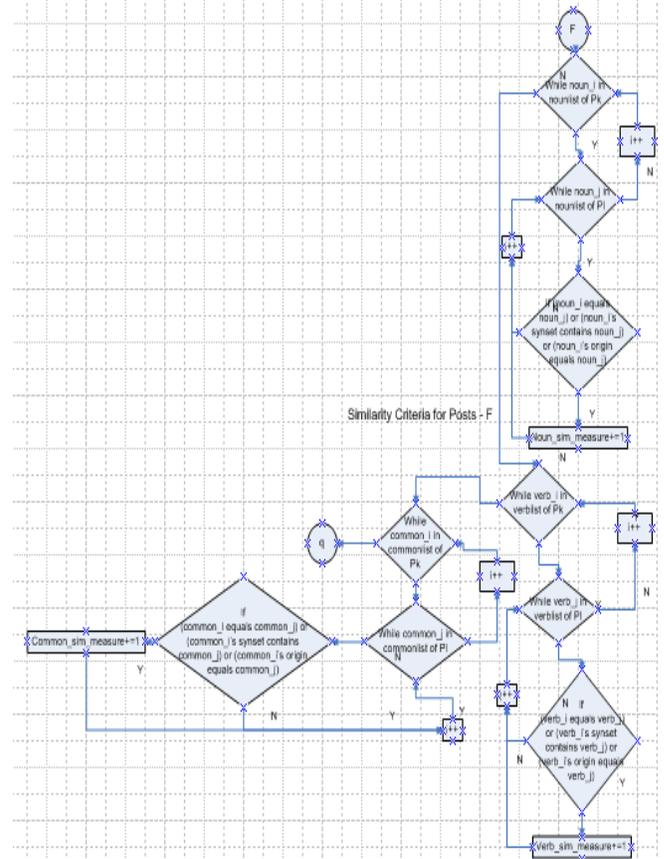

**Figure 8:** Similarity measure criteria for sentences of blog-posts

It is empirically evident that (NLP based) natural language processing based semantic similarity measure proposed in this paper is superior to tf-idf based approaches. As tf-idf based approaches only consider the term but not the kind of word of the language.

## VI. RESULT

The results obtained from the algorithm of sentence-based semantic similarity measure between blog posts is considered appropriate within the range of <0 – 1>. It means that the more similar the blog posts are; the similarity measure is more close to 1. Likewise, the least similar are the blog posts; the similarity measure is more close to 0. A graphical representation of the results of Sentence-Based Semantic Similarity Measure obtained on the testing dataset comprising of 20 blog-posts of particular political issues is shown in the Figure 9.

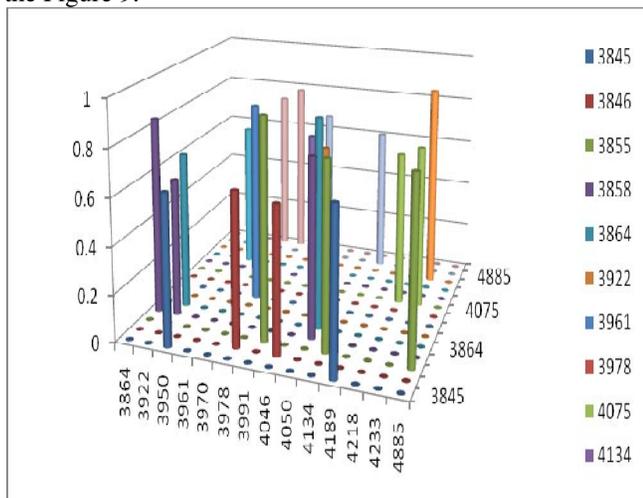

**Figure 9:** Results of Sentence-Based Semantic Similarity Measure between blog posts over various political issues

It has been observed that the blog posts collected for measuring similarity belong to the same category "Pakistani Politics" and also to the same sub-category "Political Issue". Therefore, all the blog-posts compared over the same "Political Issues" have an average similarity of 70%.

## VII. FUTURE WORK

A new approach for measuring semantic similarity of two blog-posts is proposed in this paper. Semantic similarity measure of text contents are widely used in various disciplines like language modeling, word sense disambiguation, document clustering, and search filtering. It is the sentence-based algorithm that extracts the noun-phrases, verb-phrases and common-bag-of-words to compute the similarity. The approach is a mix of corpus based and dictionary based semantic mining. We propose this methodology to be further used to identify influential bloggers from blogging community. This would involve sentence-based similarity measure as one of the effective influence measuring factors utilized in an effective algorithm for deducing influential bloggers' list. As we have observed that a lot of research work is carried out on influential bloggers mining systems, but all these systems suffer from drawbacks like: domain driven, generalized shallow influential measure and validation and verification. Therefore, our effort is to create an effective algorithm using semantic measures to identify the influential bloggers from the blogging community.